%% file: codemixedParsing.tex
\documentclass[11pt,a4paper]{article}
\usepackage[hyperref]{naaclhlt2018}
\usepackage{times}
\usepackage{latexsym}
\usepackage{caption}
\usepackage{scalefnt}
\usepackage{covington}
\usepackage{color}
\usepackage{xcolor}
\usepackage{enumitem}
\usepackage{graphicx}
\usepackage{multirow}
\usepackage{amsmath}
\usepackage{amssymb}
\usepackage{latexsym}
\usepackage{url}
\usepackage{scalefnt}
\usepackage{tikz, tikz-dependency}
\usetikzlibrary{shapes,arrows}
\usepackage{pifont}

\aclfinalcopy % Uncomment this line for the final submission
\aclfinalcopy

\title{Universal Dependency Parsing for Hindi-English Code-switching}

\author{Irshad Ahmad Bhat \\
  LTRC, IIIT-H, \\ Hyderabad, India \\
  {irshad.bhat@iiit.ac.in } \\ \And 
  Riyaz Ahmad Bhat \\ 
  Interaction Labs, \\ Bangalore, India \\
  {rbhat@interactions.com  }  \\ \And
  Manish Shrivastava \\
  LTRC, IIIT-H, \\ Hyderabad, India \\
  {m.shrivastava@iiit.ac.in} \\ \And
  Dipti Misra Sharma\\
  LTRC, IIIT-H, \\ Hyderabad, India \\
  {dipti@iiit.ac.in} \\
  }

\date{}

\begin{document}
\maketitle
\begin{abstract}
Code-switching is a phenomenon of mixing grammatical structures of two or more languages under varied social constraints. The code-switching data differ so radically from the benchmark corpora used in NLP community that the application of standard technologies to these data degrades their performance sharply. Unlike standard corpora, these data often need to go through additional processes such as language identification, normalization and/or back-transliteration for their efficient processing. In this paper, we investigate these indispensable processes and other problems associated with syntactic parsing of code-switching data and propose methods to mitigate their effects. In particular, we study dependency parsing of code-switching data of Hindi and English multilingual speakers from Twitter. We present a treebank of Hindi-English code-switching tweets under Universal Dependencies scheme and propose a neural stacking model for parsing that efficiently leverages part-of-speech tag and syntactic tree annotations in the code-switching treebank and the preexisting Hindi and English treebanks. We also present normalization and back-transliteration models with a decoding process tailored for code-switching data. Results show that our neural stacking parser is 1.5\% LAS points better than the augmented parsing model and our decoding process improves results by 3.8\% LAS points over the first-best normalization and/or back-transliteration.
\end{abstract}

\section{Introduction}
Code-switching\footnote{Code-mixing is another term in the linguistics literature used interchangeably with code-switching. Both terms are often used to refer to the same or similar phenomenon of mixed language use.} (henceforth CS) is the juxtaposition, within the same speech utterance, of grammatical units such as words, phrases, and clauses belonging to two or more different languages \cite{gumperz1982discourse}. The phenomenon is prevalent in multilingual societies where speakers share more than one language and is often prompted by multiple social factors \cite{myers1995social}. Moreover, code-switching is mostly prominent in colloquial language use in daily conversations, both online and offline. 

Most of the benchmark corpora used in NLP for training and evaluation are based on edited monolingual texts which strictly adhere to the norms of a language related, for example, to orthography, morphology, and syntax. Social media data in general and CS data, in particular, deviate from these norms implicitly set forth by the choice of corpora used in the community. This is the reason why the current technologies often perform miserably on social media data, be it monolingual or mixed language data \cite{solorio2008part,vyas2014pos,ccetinouglu-schulz-vu:2016:W16-58,gimpel2011part,owoputi2013improved,kong1001dependency}. CS data offers additional challenges over the monolingual social media data as the phenomenon of code-switching transforms the data in many ways, for example, by creating new lexical forms and syntactic structures by mixing morphology and syntax of two languages making it much more diverse than any monolingual corpora \cite{ccetinouglu-schulz-vu:2016:W16-58}. As the current computational models fail to cater to the complexities of CS data, there is often a need for dedicated techniques tailored to its specific characteristics. 

Given the peculiar nature of CS data, it has been widely studied in linguistics literature \cite{poplack1980sometimes,gumperz1982discourse,myers1995social}, and more recently, there has been a surge in studies concerning CS data in NLP as well \cite[and others]{solorio2008learning,solorio2008learning,vyas2014pos,sharma-EtAl:2016:N16-1,rudra-EtAl:2016:EMNLP2016,joshi-EtAl:2016:COLING,bhat-EtAl:2017:EACLshort,chandu2017webshodh,rijhwani2017estimating,guzman2017metrics}. Besides the individual computational works, a series of shared-tasks and workshops on preprocessing and shallow syntactic analysis of CS data have also been conducted at multiple venues such as Empirical Methods in NLP (EMNLP 2014 and 2016), International Conference on NLP (ICON 2015 and 2016) and Forum for Information Retrieval Evaluation (FIRE 2015 and 2016). Most of these works have attempted to address preliminary tasks such as language identification, normalization and/or back-transliteration as these data often need to go through these additional processes for their efficient processing. In this paper, we investigate these indispensable processes and other problems associated with syntactic parsing of code-switching data and propose methods to mitigate their effects. In particular, we study dependency parsing of Hindi-English code-switching data of multilingual Indian speakers from Twitter. Hindi-English code-switching presents an interesting scenario for the parsing community. Mixing among typologically diverse languages will intensify structural variations which will make parsing more challenging. For example, there will be many sentences containing: (1) both SOV and SVO word orders\footnote{Order of Subject, Object and Verb in transitive sentences.}, (2) both head-initial and head-final genitives, (3) both prepositional and postpositional phrases, etc. More importantly, none among the Hindi and English treebanks would provide any training instance for these mixed structures within individual sentences. In this paper, we present the first code-switching treebank that provides syntactic annotations required for parsing mixed-grammar syntactic structures. Moreover, we present a parsing pipeline designed explicitly for Hindi-English CS data. The pipeline comprises of several modules such as a language identification system, a back-transliteration system, and a dependency parser. The gist of these modules and our overall research contributions are listed as follows:

\begin{itemize}
\item back-transliteration and normalization models based on encoder-decoder frameworks with sentence decoding tailored for code-switching data; 
\item a dependency treebank of Hindi-English code-switching tweets under Universal Dependencies scheme; and 
\item a neural parsing model which learns POS tagging and parsing jointly and also incorporates knowledge from the monolingual treebanks using neural stacking.
\end{itemize}

\section{Preliminary Tasks}
\label{sec:preproc}
As preliminary steps before parsing of CS data, we need to identify the language of tokens and normalize and/or back-transliterate them to enhance the parsing performance. These steps are indispensable for processing CS data and without them the performance drops drastically as we will see in Results Section. We need normalization of non-standard word forms and back-transliteration of Romanized Hindi words for addressing out-of-vocabulary problem, and lexical and syntactic ambiguity introduced due to contracted word forms. As we will train separate normalization and back-transliteration models for Hindi and English, we need language identification for selecting which model to use for inference for each word form separately. Moreover, we also need language information for decoding best word sequences.

\subsection{Language Identification}\enspace For language identification task, we train a multilayer perceptron (MLP) stacked on top of a recurrent bidirectional LSTM (Bi-LSTM) network as shown in Figure \ref{fig:lid_model}. 

\vspace{0.5em}
\noindent
\begin{minipage}{0.95\columnwidth}
\begin{center}
\resizebox{!}{.85\columnwidth}{\includegraphics{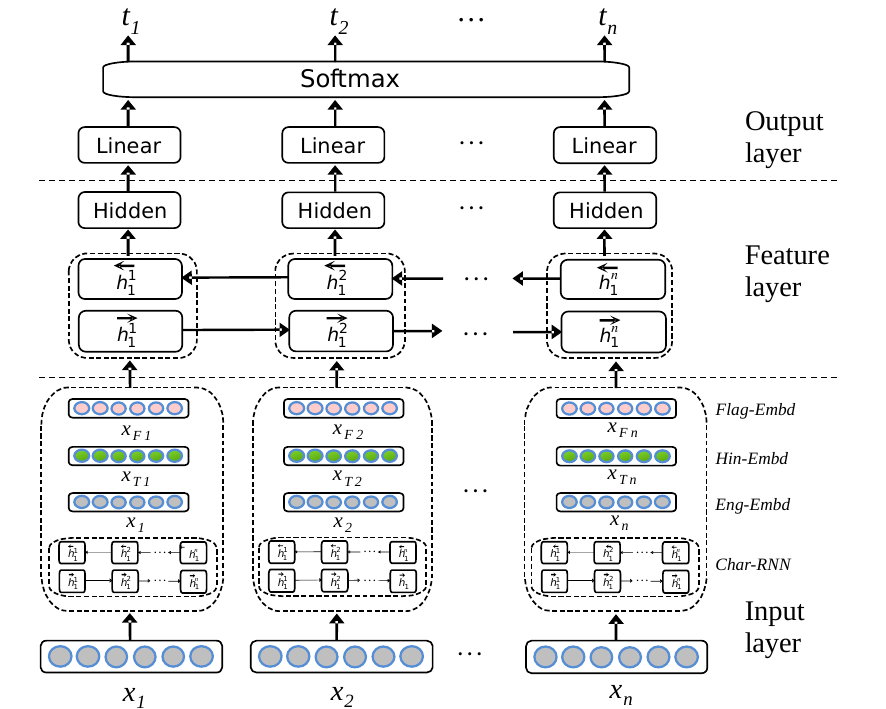}}
\captionsetup{skip=0.5em}
\captionof{figure}{Language identification network}
\label{fig:lid_model}
\end{center}
\end{minipage}
\vspace{0.25em}

\noindent We represent each token by a concatenated vector of its English embedding, back-transliterated Hindi embedding, character Bi-LSTM embedding and flag embedding (English dictionary flag and word length flag with length bins of 0-3, 4-6, 7-10, and 10-all). These concatenated vectors are passed to a Bi-LSTM network to generate a sequence of hidden representations which encode the contextual information spread across the sentence. Finally, output layer uses the feed-forward neural network with a softmax function for a probability distribution over the language tags. We train the network on our CS training set concatenated with the data set provided in ICON 2015\footnote{http://ltrc.iiit.ac.in/icon2015/} shared task (728 Facebook comments) on language identification and evaluate it on the datasets from \newcite{bhat-EtAl:2017:EACLshort}. We achieved the state-of-the-art performance on both development and test sets \cite{bhat-EtAl:2017:EACLshort}. The results are shown in Table \ref{tbl:lidResults}.

\vspace{.75em}
\noindent
\begin{minipage}{\linewidth}
%\begin{table}
\begin{center}
\resizebox{\columnwidth}{!}{%
\input{ntables/lid_acc.tex}}
%\captionsetup{font=scriptsize}
\captionsetup{skip=0.5em}
\captionof{table}{Language Identification results on CS test set.}
\label{tbl:lidResults}
\end{center}
%\end{table}
\end{minipage}

\subsection{Normalization and Back-transliteration}\enspace We learn two separate but similar character-level models for normalization-cum-transliteration of noisy Romanized Hindi words and normalization of noisy English words. We treat both normalization and back-transliteration problems as a general sequence to sequence learning problem. In general, our goal is to learn a mapping for non-standard English and Romanized Hindi word forms to standard forms in their respective scripts. In case of Hindi, we address the problem of normalization and back-transliteration of Romanized Hindi words using a single model. We use the attention-based encoder-decoder model of Luong \cite{luong2015effective} with global attention for learning. For Hindi, we train the model on the transliteration pairs (87,520) from the Libindic transliteration project\footnote{https://github.com/libindic/indic-trans} and Brahmi-Net \cite{kunchukuttan2015brahmi} which are further augmented with noisy transliteration pairs (1,75,668) for normalization. Similarly, for normalization of noisy English words, we train the model on noisy word forms (4,29,715) synthetically generated from the English vocabulary. We use simple rules such as dropping non-initial vowels and replacing consonants based on their phonological proximity to generate synthetic data for normalization. Figure \ref{fig:synthetic_trans_pairs} shows some of the noisy forms generated from standard word forms using simple and finite rules which include vowel elision ({\tt please} $\rightarrow$ {\tt pls}), interchanging similar consonants and vowels ({\tt cousin} $\rightarrow$ {\tt couzin}), replacing consonant or vowel clusters with a single letter ({\tt Twitter} $\rightarrow$ {\tt Twiter}), etc. From here onwards, we will refer to both normalization and back-transliteration as normalization.

\vspace{.75em}
\hspace{-1em}
\begin{minipage}{0.95\columnwidth}
\begin{center}
\resizebox{!}{.7\columnwidth}{\includegraphics{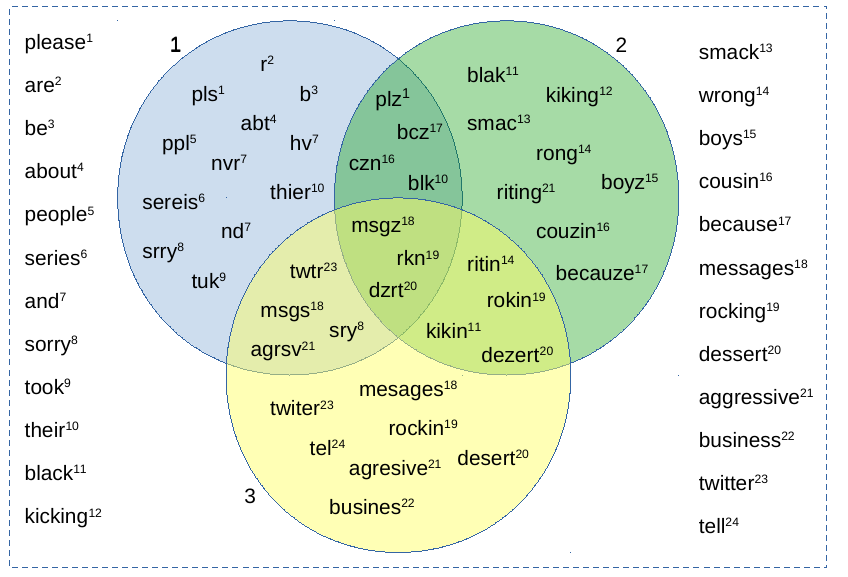}}
\captionof{figure}{Synthetic normalization pairs generated for a sample of English words using hand crafted rules.}
\label{fig:synthetic_trans_pairs}
\end{center}
\end{minipage}
%\vspace{.5em}

\begin{figure*}
\begin{center}
\resizebox{.95\linewidth}{!}{\includegraphics{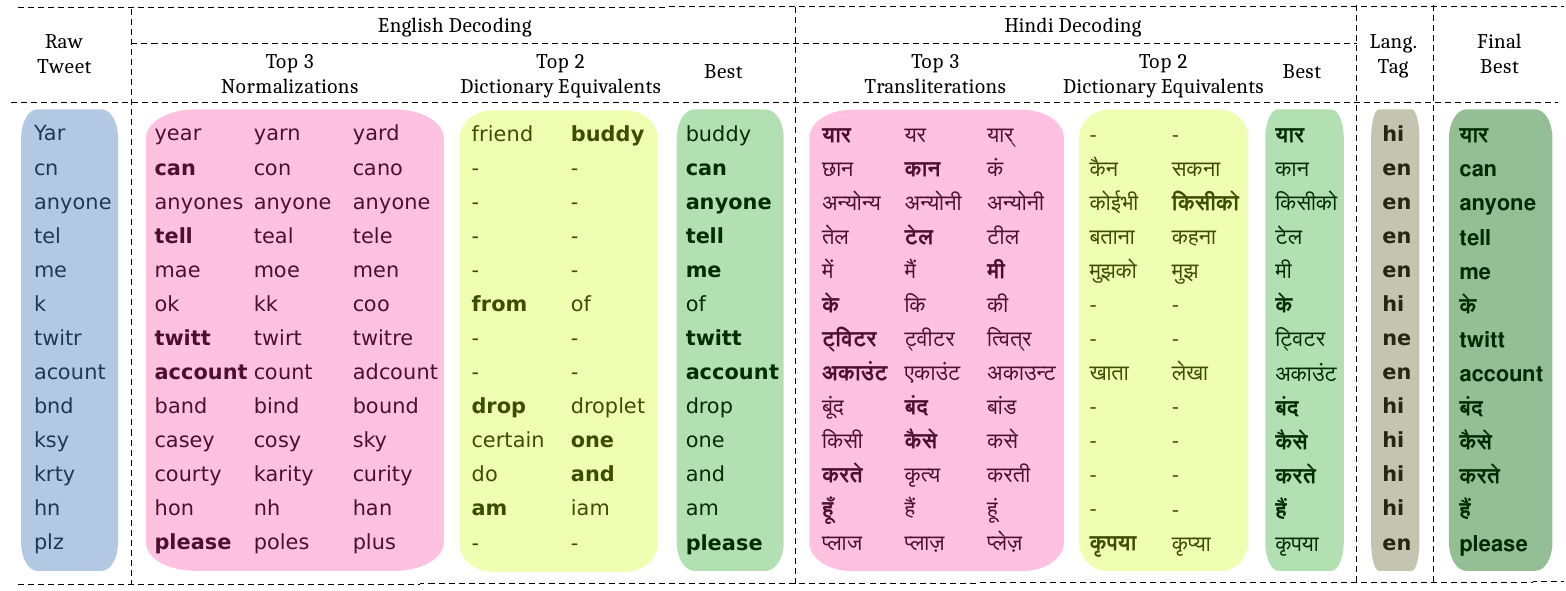}}
\captionsetup{skip=0.5em}
\captionof{figure}{The figure shows a 3-step decoding process for the sentence {\it ``Yar cn anyone tel me k twitr account bnd ksy krty hn plz''} {\it(Friend can anyone tell me how to close twitter account please)}.}
\label{fig:viterbi_trans}
\end{center}
\end{figure*}

At inference time, our normalization models will predict the most likely word form for each input word. However, the single-best output from the model may not always be the best option considering an overall sentential context. Contracted word forms in social media content are quite often ambiguous and can represent different standard word forms. For example, noisy form `{\tt pt}' can expand to different standard word forms such as `{\tt put}', `{\tt pit}', `{\tt pat}', `{\tt pot}' and `{\tt pet}'. The choice of word selection will solely depend on the sentential context. To select contextually relevant forms, we use exact search over n-best normalizations from the respective models extracted using beam-search decoding. The best word sequence is selected using the Viterbi decoding over $b^n$ word sequences scored by a trigram language model. $b$ is the size of beam-width and $n$ is the sentence length. The language models are trained on the monolingual data of Hindi and English using KenLM toolkit \cite{heafield2013scalable}. For each word, we extract five best normalizations ($b$=5). Decoding the best word sequence is a non-trivial problem for CS data due to lack of normalized and back-transliterated CS data for training a language model. One obvious solution is to apply decoding on individual language fragments in a CS sentence \cite{dutta2015text}. One major problem with this approach is that the language models used for scoring are trained on complete sentences but are applied on sentence fragments. Scoring individual CS fragments might often lead to wrong word selection due to incomplete context, particularly at fragment peripheries. We solve this problem by using a 3-step decoding process that works on two separate versions of a CS sentence, one in Hindi, and one in English. In the first step, we replace first-best back-transliterated forms of Hindi words by their translation equivalents using a Hindi-English bilingual lexicon.\footnote{An off-the-shelf MT system would have been appropriate for this task, however, we would first need to adapt it to CS data which in itself is a non-trivial task.} An exact search is used over the top `$5$' normalizations of English words, the translation equivalents of Hindi words and the actual word itself. In the second step, we decode best word sequence over Hindi version of the sentence by replacing best English word forms decoded from the first step by their translation equivalents. An exact search is used over the top `$5$' normalizations of Hindi words, the dictionary equivalents of decoded English words and the original words. In the final step, English and Hindi words are selected from their respective decoded sequences using the predicted language tags from the language identification system. Note that the bilingual mappings are only used to aid the decoding process by making the CS sentences lexically monolingual so that the monolingual language models could be used for scoring. They are not used in the final decoded output. The overall decoding process is shown in Figure \ref{fig:viterbi_trans}. 

Both of our normalization and back-transliteration systems are evaluated on the evaluation set of \newcite{bhat-EtAl:2017:EACLshort}. Results of our systems are reported in Table \ref{tbl:transResults} with a comparison of accuracies based on the nature of decoding used. The results clearly show the significance of our 3-step decoding over first-best and fragment-wise decoding.

\vspace{.5em}
\noindent
\begin{minipage}{\columnwidth}
%\begin{table}
\begin{center}
\resizebox{7.5cm}{!}{%
\input{ntables/trans_acc.tex}}
%\captionsetup{font=scriptsize}
\captionsetup{skip=0.5em}
\captionof{table}{Normalization accuracy based on the number of noisy tokens in the evaluation set. FB = First Best, and FW = Fragment Wise}
\label{tbl:transResults}
\end{center}
%\end{table}
\end{minipage}

\section{Universal Dependencies for Hindi-English}
Recently \newcite{bhat-EtAl:2017:EACLshort} provided a CS dataset for the evaluation of their parsing models which they trained on the Hindi and English Universal Dependency (UD) treebanks. We extend this dataset by annotating 1,448 more sentences.  Following \newcite{bhat-EtAl:2017:EACLshort} we first sampled CS data from a large set of tweets of Indian language users that we crawled from Twitter using Tweepy\footnote{http://www.tweepy.org/}--a Twitter API wrapper. We then used a language identification system trained on ICON dataset (see Section \ref{sec:preproc}) to filter Hindi-English CS tweets from the crawled Twitter data. Only those tweets were selected that satisfied a minimum ratio of 30:70(\%) code-switching. From this dataset, we manually selected 1,448 tweets for annotation. The selected tweets are thoroughly checked for code-switching ratio. For POS tagging and dependency annotation, we used Version 2 of Universal dependency guidelines \cite{de2014universal}, while language tags are assigned based on the tag set defined in \cite{codeswitch,jamatia2015part}. The dataset was annotated by two expert annotators who have been associated with annotation projects involving syntactic annotations for around 10 years. Nonetheless, we also ensured the quality of the manual annotations by carrying an inter-annotator agreement analysis. We randomly selected a dataset of 150 tweets which were annotated by both annotators for both POS tagging and dependency structures. The inter-annotator agreement has a 96.20\% accuracy for POS tagging and a 95.94\% UAS and a 92.65\% LAS for dependency parsing.

We use our dataset for training while the development and evaluation sets from \newcite{bhat-EtAl:2017:EACLshort} are used for tuning and evaluation of our models. Since the annotations in these datasets follow version 1.4 of the UD guidelines, we converted them to version 2 by using carefully designed rules. The statistics about the data are given in Table \ref{tbl:transResults}.

%\paragraph{Syntactic Peculiarities}
\vspace{0.5em}
\noindent
\begin{minipage}{0.99\linewidth}
%\begin{table}
\begin{center}
\resizebox{7.8cm}{!}{%
\input{ntables/data_statistics.tex}}
%\captionsetup{font=scriptsize}
\captionsetup{skip=0.5em}
\captionof{table}{Data Statistics. Dev set is used for tuning model parameters, while Test set is used for evaluation.}
\label{tbl:transResults}
\end{center}
%\end{table}
\end{minipage}

\section{Dependency Parsing}
We adapt Kiperwasser and Goldberg \shortcite{kiperwasser2016simple} transition-based parser as our base model and incorporate POS tag and monolingual parse tree information into the model using neural stacking, as shown in Figures \ref{fig:stack-prop} and \ref{fig:snnparser}. 

\subsection{Parsing Algorithm}
Our parsing models are based on an arc-eager transition system \cite{nivre2003efficient}. The arc-eager system defines a set of configurations for a sentence {\tt \small w$_1$,...,w$_n$}, where each configuration {\tt \small C = (S, B, A)} consists of a stack {\tt \small S}, a buffer {\tt \small B}, and a set of dependency arcs {\tt \small A}. For each sentence, the parser starts with an initial configuration where {\tt \small S = [ROOT], B = [w$_1$,...,w$_n$]} and {\tt \small A = $\emptyset$} and terminates with a configuration {\tt \small C} if the buffer is empty and the stack contains the {\tt \small ROOT}. The parse trees derived from transition sequences are given by {\tt \small A}. To derive the parse tree, the arc-eager system defines four types of transitions ({\tt \small $t$}): {\tt \small Shift}, {\tt \small Left-Arc}, {\tt \small Right-Arc}, and {\tt \small Reduce}.

We use the training by exploration method of \newcite{goldberg2012dynamic} for decoding a transition sequence which helps in mitigating error propagation at evaluation time. We also use pseudo-projective transformations of \newcite{nivre2005} to handle a higher percentage of non-projective arcs in the CS data ($\sim$2\%). We use the most informative scheme of {\tt head+path} to store the transformation information.

\vspace{1em}
\noindent
\begin{minipage}{0.95\linewidth}
\begin{center}
\resizebox{\columnwidth}{!}{\includegraphics{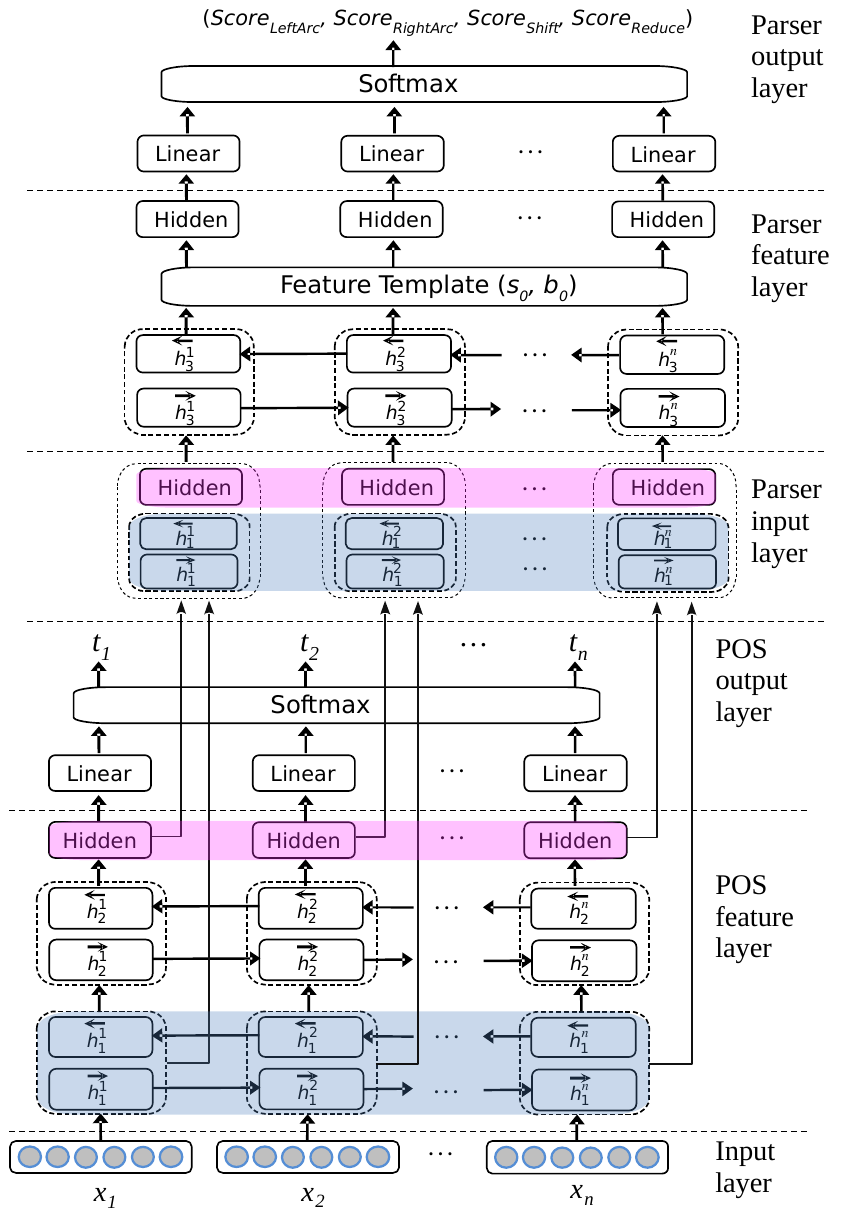}}
\captionsetup{skip=0.5em}
\captionof{figure}{POS tagging and parsing network based on stack-propagation model proposed in \cite{zhang-weiss:2016:P16-1}.}
\label{fig:stack-prop}
\end{center}
\end{minipage}

\subsection{Base Models}
Our base model is a stack of a tagger network and a parser network inspired by stack-propagation model of \newcite{zhang-weiss:2016:P16-1}. The parameters of the tagger network are shared and act as a regularization on the parsing model. The model is trained by minimizing a joint negative log-likelihood loss for both tasks. Unlike \newcite{zhang-weiss:2016:P16-1}, we compute the gradients of the log-loss function simultaneously for each training instance. While the parser network is updated given the parsing loss only, the tagger network is updated with respect to both tagging and parsing losses. Both tagger and parser networks comprise of an input layer, a feature layer, and an output layer as shown in Figure \ref{fig:stack-prop}. Following \newcite{zhang-weiss:2016:P16-1}, we refer to this model as stack-prop.

\paragraph{Tagger network:} The input layer of the tagger encodes each input word in a sentence by concatenating a pre-trained word embedding with its character embedding given by a character Bi-LSTM. In the feature layer, the concatenated word and character representations are passed through two stacked Bi-LSTMs to generate a sequence of hidden representations which encode the contextual information spread across the sentence. The first Bi-LSTM is shared with the parser network while the other is specific to the tagger. Finally, output layer uses the feed-forward neural network with a softmax function for a probability distribution over the Universal POS tags. We only use the forward and backward hidden representations of the focus word for classification.

\paragraph{Parser Network:} Similar to the tagger network, the input layer encodes the input sentence using word and character embeddings which are then passed to the shared Bi-LSTM. The hidden representations from the shared Bi-LSTM are then concatenated with the dense representations from the feed-forward network of the tagger and passed through the Bi-LSTM specific to the parser. This ensures that the tagging network is penalized for the parsing error caused by error propagation by back-propagating the gradients to the shared tagger parameters \cite{zhang-weiss:2016:P16-1}. Finally, we use a non-linear feed-forward network to predict the labeled transitions for the parser configurations. From each parser configuration, we extract the top node in the stack and the first node in the buffer and use their hidden representations from the parser specific Bi-LSTM for classification.

%\vspace{-1em}
\begin{center}
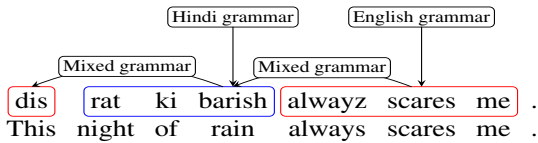

\resizebox{7.5cm}{.25\columnwidth}{\input{nfigures/cdexample.tex}}
\captionsetup{skip=0.5em}
\captionof{figure}{Code-switching tweet showing grammatical fragments from Hindi and English.}
\label{tree:cdexample}
\end{center}

\subsection{Stacking Models}
It seems reasonable that limited CS data would complement large monolingual data in parsing CS data and a parsing model which leverages both data would significantly improve parsing performance. While a parsing model trained on our limited CS data might not be enough to accurately parse the individual grammatical fragments of Hindi and English, the preexisting Hindi and English treebanks are large enough to provide sufficient annotations to capture their structure. Similarly, parsing model(s) trained on the Hindi and English data may not be able to properly connect the divergent fragments of the two languages as the model lacks evidence for such mixed structures in the monolingual data. This will happen quite often as Hindi and English are typologicalls very diverse (see Figure \ref{tree:cdexample}). 

\vspace{.5em}
\hspace{-1em}
\noindent
\begin{minipage}{\linewidth}
\begin{center}
\resizebox{1.1\columnwidth}{!}{\includegraphics{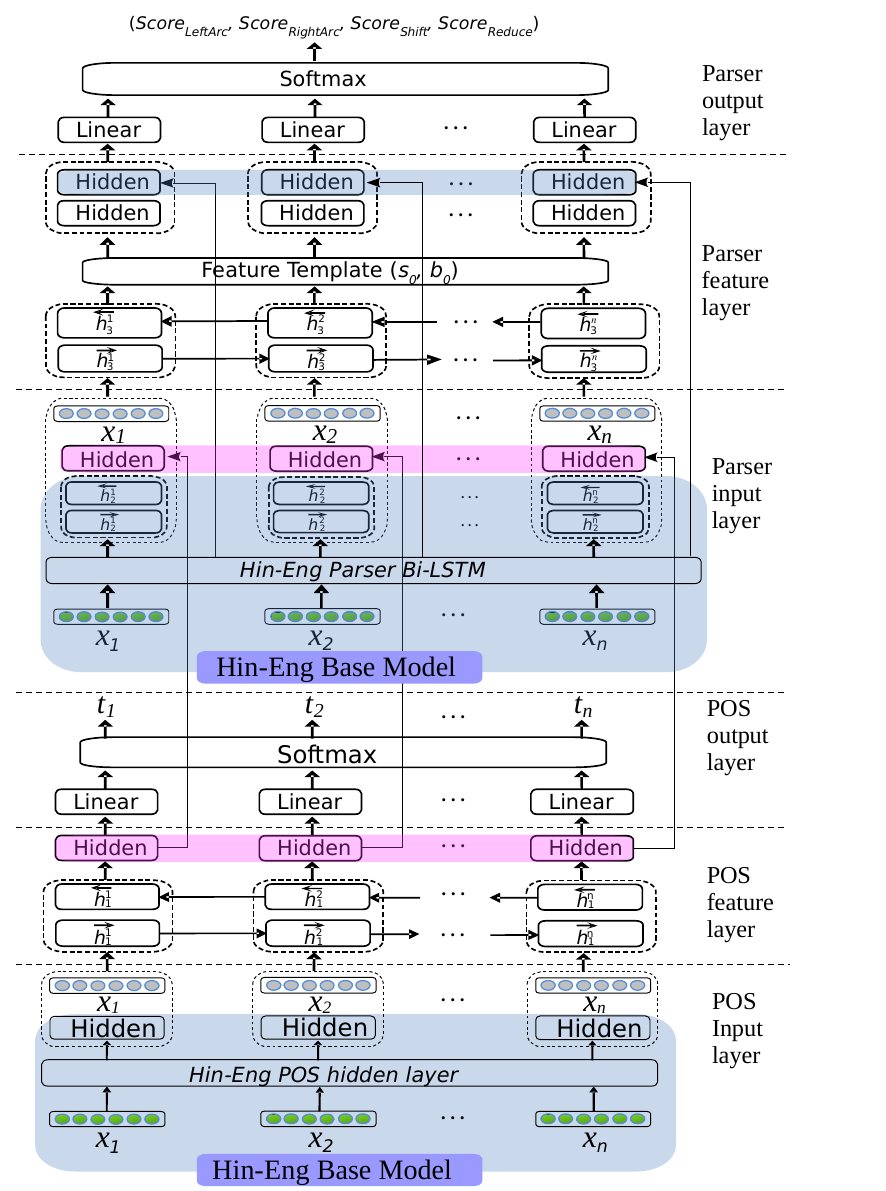}}
\captionsetup{skip=0.5em}
\captionof{figure}{Neural Stacking-based parsing architecture for incorporating monolingual syntactic knowledge.}
\label{fig:snnparser}
\end{center}
\end{minipage}
\vspace{.5em}

As we discussed above, we adapted feature-level neural stacking \cite{zhang-weiss:2016:P16-1,chen-zhang-liu:2016:EMNLP2016} for joint learning of POS tagging and parsing. Similarly, we also adapt this stacking approach for incorporating the monolingual syntactic knowledge into the base CS model. Recently, \newcite{wang-EtAl:2017:Long6} used neural stacking for injecting syntactic knowledge of English into a graph-based Singlish parser which lead to significant improvements in parsing performance. Unlike \newcite{wang-EtAl:2017:Long6}, our base stacked models will allow us to transfer the POS tagging knowledge as well along the parse tree knowledge.

As shown in Figure \ref{fig:snnparser}, we transfer both POS tagging and parsing information from the source model trained on augmented Hindi and English data. For tagging, we augment the input layer of the CS tagger with the MLP layer of the source tagger. For transferring parsing knowledge, hidden representations from the parser specific Bi-LSTM of the source parser are augmented with the input layer of the CS parser which already includes the hidden layer of the CS tagger, word and character embeddings. In addition, we also add the MLP layer of the source parser to the MLP layer of the CS parser. The MLP layers of the source parser are generated using raw features from CS parser configurations. Apart from the addition of these learned representations from the source model, the overall CS model remains similar to the base model shown in Figure \ref{fig:stack-prop}. The tagging and parsing losses are back-propagated by traversing back the forward paths to all trainable parameters in the entire network for training and the whole network is used collectively for inference.

\section{Experiments}
\label{sec:expsetup}
We train all of our POS tagging and parsing models on training sets of the Hindi and English UD-v2 treebanks and our Hindi-English CS treebank. For tuning and evaluation, we use the development and evaluation sets from \newcite{bhat-EtAl:2017:EACLshort}. We conduct multiple experiments in gold and predicted settings to measure the effectiveness of the sub-modules of our parsing pipeline. In predicted settings, we use the POS taggers separately trained on the Hindi, English and CS training sets. All of our models use word embeddings from transformed Hindi and English embedding spaces to address the problem of lexical differences prevalent in CS sentences.

\subsection{Hyperparameters}
\paragraph{Word Representations}\enspace For language identification, POS tagging and parsing models, we include the lexical features in the input layer of our neural networks using 64-dimension pre-trained word embeddings, while we use randomly initialized embeddings within a range of $[-0.1$, $+0.1]$ for non-lexical units such as POS tags and dictionary flags. We use 32-dimensional character embeddings for all the three models and 32-dimensional POS tag embeddings for pipelined parsing models. The distributed representation of Hindi and English vocabulary are learned separately from the Hindi and English monolingual corpora. The English monolingual data contains around 280M sentences, while the Hindi data is comparatively smaller and contains around 40M sentences. The word representations are learned using Skip-gram model with negative sampling which is implemented in {\tt word2vec} toolkit \cite{mikolov2013efficient}. We use the projection algorithm of \newcite{artetxe2016learning} to transform the Hindi and English monolingual embeddings into same semantic space using a bilingual lexicon ($\sim$63,000 entries). The bilingual lexicon is extracted from ILCI and Bojar Hindi-English parallel corpora \cite{jha2010tdil,hindencorp05:lrec:2014}. For normalization models, we use 32-dimensional character embeddings uniformly initialized within a range of $[-0.1, +0.1]$.

\paragraph{Hidden dimensions} The POS tagger specific Bi-LSTMs have 128 cells while the parser specific Bi-LSTMs have 256 cells. The Bi-LSTM in the language identification model has 64 cells. The character Bi-LSTMs have 32 cells for all three models. The hidden layer of MLP has 64 nodes for the language identification network, 128 nodes for the POS tagger and 256 nodes for the parser. We use hyperbolic tangent as an activation function in all tasks. In the normalization models, we use single layered Bi-LSTMs with 512 cells for both encoding and decoding of character sequences.

\paragraph{Learning} For language identification, POS tagging and parsing networks, we use momentum SGD for learning with a minibatch size of 1. The LSTM weights are initialized with random orthonormal matrices as described in \cite{saxe2013exact}. We set the dropout rate to 30\% for POS tagger and parser Bi-LSTM and MLP hidden states while for language identification network we set the dropout to 50\%. All three models are trained for up to 100 epochs, with early stopping based on the development set.

In case of normalization, we train our encoder-decoder models for 25 epochs using vanilla SGD. We start with a learning rate of $1.0$ and after 8 epochs reduce it to half for every epoch. We use a mini-batch size of 128, and the normalized gradient is rescaled whenever its norm exceeds 5. We use a dropout rate of 30\% for the Bi-LSTM.

Language identification, POS tagging and parsing code is implemented in DyNet \cite{neubig2017dynet} and for normalization without decoding, we use Open-NMT toolkit for neural machine translation \cite{opennmt}. All the code is available at \url{https://github.com/irshadbhat/nsdp-cs} and the data is available at \url{https://github.com/CodeMixedUniversalDependencies/UD_Hindi_English}.

\section{Results}
In Table \ref{tbl:jointparsing}, we present the results of our main model that uses neural stacking for learning POS tagging and parsing and also for knowledge transfer from the Bilingual model. Transferring POS tagging and syntactic knowledge using neural stacking gives 1.5\% LAS\footnote{The improvements discussed in the running text are for the models that are evaluated in auto settings.} improvement over a naive approach of data augmentation. The Bilingual model which is trained on the union of Hindi and English data sets is least accurate of all our parsing models. However, it achieves better or near state-of-the-art results on the Hindi and English evaluation sets (see Table \ref{tbl:nocmJointModelResults}). As compared to the best system in CoNLL 2017 Shared Task on Universal Dependencies \cite{zeman-EtAl:2017:K17-3,dozat2017stanford}, our results for English are around 3\% better in LAS, while for Hindi only 0.5\% LAS points worse. The CS model trained only on the CS training data is slightly more accurate than the Bilingual model. Augmenting the CS data to Hindi-English data complements their syntactic structures relevant for parsing mixed grammar structures which are otherwise missing in the individual datasets. The average improvements of around $\sim$5\% LAS clearly show their complementary nature.

\vspace{.75em}
\noindent
\begin{minipage}{\linewidth}
\begin{center}
\resizebox{\columnwidth}{!}{\input{ntables/jointmodel_las.tex}}
%\captionsetup{font=scriptsize}
\captionsetup{skip=0.5em}
\captionof{table}{Accuracy of different parsing models on the evaluation set. POS tags are jointly predicted with parsing. LID = Language tag, TRN = Transliteration/normalization.}
\label{tbl:jointparsing}
\end{center}
\end{minipage}
\vspace{.5em}

Table \ref{tbl:posResults} summarizes the POS tagging results on the CS evaluation set. The tagger trained on the CS training data is 2.5\% better than the Bilingual tagger. Adding CS training data to Hindi and English train sets further improves the accuracy by 1\%. However, our stack-prop tagger achieves the highest accuracy of 90.53\% by leveraging POS information from Bilingual tagger using neural stacking.

\vspace{.7em}
\noindent
\begin{minipage}{\linewidth}
\begin{center}
\resizebox{\columnwidth}{!}{\input{ntables/no_cm_jointmodel_las.tex}}
%\captionsetup{font=scriptsize}
\captionsetup{skip=0.5em}
\captionof{table}{POS and parsing results for Hindi and English monolingual test sets using pipeline and stack-prop models.}
\label{tbl:nocmJointModelResults} 
\end{center}
\end{minipage}

\vspace{.7em}
\noindent
\begin{minipage}{\linewidth}
\begin{center}
\resizebox{\columnwidth}{!}{%
\input{ntables/pos_acc.tex}}
%\captionsetup{font=scriptsize}
\captionsetup{skip=0.5em}
\captionof{table}{POS tagging accuracies of different models on CS evaluation set. SP = stack-prop.}
\label{tbl:posResults} 
\end{center}
\end{minipage}

\paragraph{Pipeline vs Stack-prop}
Table \ref{tbl:pipelineResults} summarizes the parsing results of our pipeline models which use predicted POS tags as input features. As compared to our stack-prop models (Table \ref{tbl:jointparsing}), pipeline models are less accurate (average 1\% LAS improvement across models) which clearly emphasizes the significance of back-propagating the parsing loss to tagging parameters as well. %For pipeline-based Bilingual and CS parsers, we use the predicted tags from the stack-prop tagger which is the reason why there are more accurate than the stack-prop parsing models. 

\vspace{.7em}
\noindent
\begin{minipage}{\linewidth}
\begin{center}
\resizebox{\columnwidth}{!}{\input{ntables/pipeline_las.tex}}
%\captionsetup{font=scriptsize}
\captionsetup{skip=0.5em}
\captionof{table}{Accuracy of different parsing models on the test set using predicted language tags, normalized/back-transliterated words and predicted POS tags. POS tags are predicted separately before parsing. In Neural Stacking model, only parsing knowledge from the Bilingual model is transferred.}
\label{tbl:pipelineResults} 
\end{center}
\end{minipage}

\paragraph{Significance of normalization} We also conducted experiments to evaluate the impact of normalization on both POS tagging and parsing. The results are shown in Table \ref{tbl:normonparse}. As expected, tagging and parsing models that use normalization without decoding achieve an average of 1\%  improvement over the models that do not use normalization at all. However, our 3-step decoding leads to higher gains in tagging as well as parsing accuracies. We achieved around 2.8\% improvements in tagging and around 4.6\% in parsing over the models that use first-best word forms from the normalization models. More importantly, there is a moderate drop in accuracy (1.4\% LAS points) caused due to normalization errors (see results in Table \ref{tbl:jointparsing} for gold vs auto normalization).

\vspace{.7em}
\noindent
\begin{minipage}{\linewidth}
\begin{center}
\resizebox{\columnwidth}{!}{%
\input{ntables/notrans_vs_1stbest_vs_viterbi.tex}}
%\captionsetup{font=scriptsize}
\captionsetup{skip=0.5em}
\captionof{table}{Impact of normalization and back-transliteration on POS tagging and parsing models.}
\label{tbl:normonparse}
\end{center}
\end{minipage}

\paragraph{Monolingual vs Cross-lingual Embeddings} We also conducted experiments with monolingual and cross-lingual embeddings to evaluate the need for transforming the monolingual embeddings into a same semantic space for processing of CS data. Results are shown in Table \ref{tbl:monoVsCL}. Cross-lingual embeddings have brought around $\sim$0.5\% improvements in both tagging and parsing. Cross-lingual embeddings are essential for removing lexical differences which is one of the problems encountered in CS data. Addressing the lexical differences will help in better learning by exposing syntactic similarities between languages.

\vspace{.7em}
\noindent
\begin{minipage}{\linewidth}
\begin{center}
\resizebox{\columnwidth}{!}{%
\input{ntables/mono_vs_cl_embd.tex}}
\captionsetup{skip=0.5em}
\captionof{table}{Impact of monolingual and cross-lingual embeddings on stacking model performance.}
\label{tbl:monoVsCL}
\end{center}
\end{minipage}

\section{Conclusion}
In this paper, we have presented a dependency parser designed explicitly for Hindi-English CS data. The parser uses neural stacking architecture of \newcite{zhang-weiss:2016:P16-1} and \newcite{chen-zhang-liu:2016:EMNLP2016} for learning POS tagging and parsing and for knowledge transfer from Bilingual models trained on Hindi and English UD treebanks. We have also presented normalization and back-transliteration models with a decoding process tailored for CS data. Our neural stacking parser is 1.5\% LAS points better than the augmented parsing model and 3.8\% LAS points better than the one which uses first-best normalizations.

%The code of the parsing models is available at the GitHub repository\enspace\path{https://github.com/irshadbhat/cm-parser}, while the data can be found under the Universal Dependencies of Hindi at\enspace\pathhttps://github.com/UniversalDependencies/UD_Hindi}.

%\section*{Acknowledgments}
%Here goes the acknowledgement!

\bibliographystyle{aclNatbib}
\bibliography{naaclhlt2018}

\appendix

\clearpage
\section{Supplemental Material}
\label{sec:supplemental}

\subsection{Example Annotations from our CS Treebank}
\input{nfigures/cm_examples.tex}
%Submissions may include resources (software and/or data) used in in the work and described in the paper. Papers that are submitted with accompanying software and/or data may receive additional credit toward the overall evaluation score, and the potential impact of the software and data will be taken into account when making the acceptance/rejection decisions. Any accompanying software and/or data should include licenses and documentation of research review as appropriate.

%NAACL-HLT also encourages the submission of supplementary material to report preprocessing decisions, model parameters, and other details necessary for the replication of the experiments reported in the paper. Seemingly small preprocessing decisions can sometimes make a large difference in performance, so it is crucial to record such decisions to precisely characterize state-of-the-art methods.

%Appendices ({\em i.e.} supplementary material in the form of proofs, tables, or pseudo-code) should come after the references, as shown here. Use \verb|\appendix| before any appendix section to switch the section numbering over to letters.

%\section{Multiple Appendices}
%\dots can be gotten by using more than one section. We hope you won't need that.

\end{document}

%% file: ntables/lid_acc.tex
%\documentclass{standalone}
%\usepackage{caption}
%\usepackage{scalefnt}
%\usepackage{multicol,multirow}
%\usepackage{amssymb}
%\usepackage{amssymb}
%
%\begin{document}
%%\resizebox{1\linewidth}{!}{%
\begin{tabular}{|c|c|c|c||c|}\hline
Label   &  Precision  & Recall  & F1-Score  &  count  \\ \hline \hline
hi      &    97.76    &  98.09  &  97.92    &   1465  \\ \hline
en      &    96.87    &  98.83  &  97.84    &   1283  \\ \hline
ne      &    94.33    &  79.17  &  86.08    &    168  \\ \hline
acro    &    92.00    &  76.67  &  83.64    &     30  \\ \hline
univ    &    99.71    &   1.00  &  99.86    &    349  \\ \hline\hline
average   &    97.39    &  97.42  &  97.36    &   3295  \\ \hline
\cite{bhat-EtAl:2017:EACLshort}   &   -    &  96.10  &  -    &   -  \\ \hline
\end{tabular}%}%
%%\end{document}

%% file: ntables/trans_acc.tex
%\documentclass{standalone}
%\usepackage{caption}
%\usepackage{scalefnt}
%\usepackage{multicol,multirow}
%\usepackage{amssymb}
%\usepackage{amssymb}
%
%\begin{document}
%%\resizebox{1\linewidth}{!}{%
\begin{tabular}{|c|cccc|cccc|}\hline

\multirow{2}{*}{Data-set} & \multicolumn{4}{|c|}{Hindi} & \multicolumn{4}{|c|}{English} \\\cline{2-9}

       & Tokens &   FB      &   FW        &   3-step            & Tokens  &   FB     &   FW         &   3-step       \\ \hline \hline
Dev    & 1549   &   82.82   &   87.28     &   {\bf90.01}    &   34    &   82.35  &   {\bf88.23}      &   {\bf88.23}     \\ \hline 
Test   & 1465   &   83.54   &   88.19     &   {\bf90.64}         &   28    &   71.42  &   75.21      &   {\bf81.71}     \\ \hline
\end{tabular}%}%
%\end{document}

%% file: ntables/data_statistics.tex
%\documentclass{standalone}
%\usepackage{caption}
%\usepackage{scalefnt}
%\usepackage{multicol,multirow}
%\usepackage{amssymb}
%\usepackage{amssymb}
%
%\begin{document}
%%\resizebox{1\linewidth}{!}{%
\begin{tabular}{|c|cc|ccccc|}\hline

Data-set &  Sentences &   Tokens  &   Hi  &  En   & Ne  & Univ  & Acro \\ \cline{2-8} \hline
Train    &  1,448     &   20,203  & 8,363 & 8,270 & 698 & 2,730 & 142  \\ %\hline
Dev      &    225     &    3,411  & 1,549 & 1,300 & 151 &  379  & 32\\ %\hline
Test     &    225     &    3,295  & 1,465 & 1,283 & 168 &  349  & 30   \\ \hline
\end{tabular}%}%
%\end{document}

%% file: nfigures/cdexample.tex
\begin{dependency}[arc edge, arc angle=25]
\begin{deptext}[column sep=.1cm]
dis \& rat \& ki \& barish \& alwayz \& scares \& me \& . \\ [.2em]
This \& night \& of \& rain \& always \& scares \& me \& . \\
\end{deptext}
\depedge[edge unit distance=2ex, edge horizontal padding=4pt]{4}{1}{Mixed grammar}
\depedge[edge unit distance=2ex, edge horizontal padding=4pt]{6}{4}{Mixed grammar}
\deproot[edge unit distance=2ex]{4}{Hindi grammar}
\deproot[edge unit distance=2ex]{6}{English grammar}
\wordgroup[style={red}]{1}{1}{1}{none}
\wordgroup[style={blue}]{1}{2}{4}{none}
\wordgroup[style={red}]{1}{5}{7}{none}
\end{dependency}

%% file: ntables/jointmodel_las.tex
%\documentclass{standalone}
%\usepackage{caption}
%\usepackage{scalefnt}
%\usepackage{multicol,multirow}
%\usepackage{amssymb}
%\usepackage{amssymb}
%
%\begin{document}
%%\resizebox{1\linewidth}{!}{%
\begin{tabular}{|c|cc|cc|}\hline

\multirow{2}{*}{Model} & \multicolumn{2}{|c|}{Gold (LID+TRN)} & \multicolumn{2}{|c|}{Auto (LID+TRN)} \\\cline{2-5}

                &  UAS     &   LAS   &  UAS     &   LAS   \\ \hline \hline
Bilingual   	&  75.26   &   65.41 &  73.29   &   63.18 \\ \hline
CS   	        &  76.69   &   66.90 &  75.84   &   64.94 \\ \hline
Augmented        &  80.39   &   71.27 &  78.95   &   69.51 \\ \hline
Neural Stacking        &  {\bf81.50}   &   {\bf72.44} &  {\bf80.23}   &   {\bf71.03} \\ \hline
\cite{bhat-EtAl:2017:EACLshort}        &  74.16  & 64.11 &  66.18 & 54.40 \\ \hline
\end{tabular}%}%
%\end{document}

%% file: ntables/no_cm_jointmodel_las.tex
%\documentclass{standalone}
%\usepackage{caption}
%\usepackage{scalefnt}
%\usepackage{multicol,multirow}
%\usepackage{amssymb}
%\usepackage{amssymb}
%
%\begin{document}
%%\resizebox{1\linewidth}{!}{%
\begin{tabular}{|c|cc|ccc|ccc|}\hline
               & \multicolumn{5}{|c|}{Pipeline} & \multicolumn{3}{c|}{\multirow{2}{*}{Stack-prop}} \\\cline{2-6}

Data-set       & \multicolumn{2}{|c|}{Gold POS} & \multicolumn{3}{c|}{Auto POS} &  \multicolumn{3}{c|}{}\\\cline{2-9}
               &  UAS     &   LAS   & POS     & UAS     &   LAS   &  POS    &  UAS     &   LAS   \\ \hline \hline
Hindi          &  95.66   &   93.08 & 97.52   & 94.08   &   90.69 &  \bf97.65  &  \bf94.36   &   \bf91.02 \\ \hline
English        &  89.95   &   87.96 & 95.75   & 87.71   &   84.59 &  \bf95.80  &  \bf88.30   &   \bf85.30 \\ \hline
\end{tabular}%}%
%\end{document}

%% file: ntables/pos_acc.tex
%\documentclass{standalone}
%\usepackage{caption}
%\usepackage{scalefnt}
%\usepackage{multicol,multirow}
%\usepackage{amssymb}
%\usepackage{amssymb}
%
%\begin{document}
%%\resizebox{1\linewidth}{!}{%
\begin{tabular}{|c|cc|cc|}\hline

\multirow{2}{*}{Model} & \multicolumn{2}{|c|}{Gold (LID+TRN)} & \multicolumn{2}{|c|}{Auto (LID+TRN)} \\\cline{2-5}

             &  Pipeline   &   SP      &  Pipeline   &   SP      \\ \hline \hline
Bilingual    &  88.36      &   88.12   &  86.71      &   86.27   \\ \hline
CS           &  90.32      &   90.38   &  89.12      &   89.19   \\ \hline
Augmented    &  91.20      &   91.50   &  90.02      &   90.20   \\ \hline
Neural Stacking     &  {\bf91.76}      &   {\bf91.90}   &  {\bf90.36}      &   {\bf90.53}   \\ \hline
\cite{bhat-EtAl:2017:EACLshort}    &  \multicolumn{2}{|c|}{86.00} &  \multicolumn{2}{|c|}{85.30}  \\ \hline
\end{tabular}%}%
%%\end{document}

%% file: ntables/pipeline_las.tex
%\documentclass{standalone}
%\usepackage{caption}
%\usepackage{scalefnt}
%\usepackage{multicol,multirow}
%\usepackage{amssymb}
%\usepackage{amssymb}
%
%\begin{document}
%\resizebox{1\linewidth}{!}{%
\begin{tabular}{|c|cc|cc|}\hline

\multirow{2}{*}{Model} & \multicolumn{2}{|c|}{Gold (LID+TRN+POS)} & \multicolumn{2}{|c|}{Auto (LID+TRN+POS)} \\\cline{2-5}

              &  UAS     &   LAS   &  UAS     &   LAS   \\ \hline \hline
Bilingual       &  82.29   &   73.79 &  72.09   &   61.18 \\ \hline
CS            &  82.73   &   73.38 &  75.20   &   64.64 \\ \hline
Augmented    &  85.66   &   77.75 &  77.98   &   69.16 \\ \hline
Neural Stacking      &  {\bf86.87}   &   {\bf78.57} &  {\bf78.90}   &   {\bf69.45} \\ \hline
\end{tabular}%}%
%\end{document}

%% file: ntables/notrans_vs_1stbest_vs_viterbi.tex
\begin{tabular}{|c|c|cc|}\hline
System              &  POS    &  UAS     &   LAS   \\ \hline \hline          
                                             
No Normalization  &  86.98  &  76.25   &   66.02   \\ \hline
First Best          &  87.74  &  78.26   &   67.22   \\ \hline
3-step Decoding    &  {\bf90.53}  &  {\bf80.23}   &   {\bf71.03}   \\ \hline
\end{tabular}%}%

%% file: ntables/mono_vs_cl_embd.tex
%\documentclass{standalone}
%\usepackage{caption}
%\usepackage{scalefnt}
%\usepackage{multicol,multirow}
%\usepackage{amssymb}
%\usepackage{amssymb}
%
%\begin{document}
%%\resizebox{1\linewidth}{!}{%
\begin{tabular}{|c|c|c|c|}\hline

Embedding    &  POS     & UAS     &   LAS    \\ \hline \hline
Monolingual  &  90.07   & 79.46   &   70.53  \\ \hline
Crosslingual &  {\bf90.53}   & {\bf80.23}   &   {\bf71.03}  \\ \hline
\end{tabular}%}%
%\end{document}

%% file: nfigures/cm_examples.tex
%\documentclass[12pt]{article}
%\usepackage{tikz, tikz-dependency}

%\begin{document}
\centering
\resizebox{.9\linewidth}{!}{
\begin{dependency}[label theme = simple, edge theme = iron, label style={font=\small}]
 \begin{deptext}[column sep=1em]
  i \& thought \& mosam \& different \& hoga \& bas \& fog \& hy \\
 \end{deptext}
 \deproot{2}{ROOT}
 \depedge{2}{1}{nsubj}
 \depedge{4}{3}{nsubj}
 \depedge{2}{4}{ccomp}
 \depedge{4}{5}{cop}
 \depedge{7}{6}{advmod}
 \depedge[edge unit distance=1.75ex]{2}{7}{advcl}
 \depedge{7}{8}{cop}
\wordgroup[style={red}]{1}{1}{2}{none}
\wordgroup[style={blue}]{1}{3}{3}{none}
\wordgroup[style={red}]{1}{4}{4}{none}
\wordgroup[style={blue}]{1}{5}{6}{none}
\wordgroup[style={red}]{1}{7}{7}{none}
\wordgroup[style={blue}]{1}{8}{8}{none}
\end{dependency}}

\centering
\resizebox{1\linewidth}{!}{
\begin{dependency}[label theme = simple, edge theme = iron, label style={font=\small}]
 \begin{deptext}[column sep=1em]
  Thand \& bhi \& odd \& even \& formula \& follow \& Kr \& rhi \& h \& ;-) \\
 \end{deptext}
 \deproot{7}{ROOT}
 \depedge[edge unit distance=1.5ex]{7}{1}{nsubj}
 \depedge{1}{2}{advmod}
 \depedge{5}{3}{amod}
 \depedge{3}{4}{compound}
 \depedge{7}{5}{obj}
 \depedge{7}{6}{compound}
 \depedge{7}{8}{aux}
 \depedge{7}{9}{aux}
 \depedge{7}{10}{discourse}
\wordgroup[style={blue}]{1}{1}{2}{none}
\wordgroup[style={red}]{1}{3}{6}{none}
\wordgroup[style={blue}]{1}{7}{9}{none}
\end{dependency}}

%\resizebox{.7\linewidth}{!}{
%\begin{dependency}[label theme = simple, edge theme = iron, label style={font=\small}]
% \begin{deptext}[column sep=1em]
%  Sunday \& is \& the\& weekly \& ghar \& ka \& Saaf \& Safai \& day \& ! \\
% \end{deptext}
% \deproot[edge unit distance=3.75ex]{9}{ROOT}
% \depedge[edge unit distance=1.75ex]{9}{1}{nsubj}
% \depedge[edge unit distance=1.75ex]{9}{2}{cop}
% \depedge[edge unit distance=1.75ex]{9}{3}{det}
% \depedge[edge unit distance=1.75ex]{9}{4}{amod}
% \depedge[edge unit distance=1.75ex]{8}{5}{nmod}
% \depedge{5}{6}{case}
% \depedge{8}{7}{compound}
% \depedge{9}{8}{nmod}
% \depedge{9}{10}{punct}
%\end{dependency}}

\centering
\resizebox{.7\linewidth}{!}{
\begin{dependency}[label theme = simple, edge theme = iron, label style={font=\small}]
 \begin{deptext}[column sep=1em]
  Tum \& kitne \& fake \& account \& banaogy \\
 \end{deptext}
 \deproot[edge unit distance=2.5ex]{5}{ROOT}
 \depedge[edge unit distance=2ex]{5}{1}{nsubj}
 \depedge[edge unit distance=2.5ex]{4}{2}{det}
 \depedge{4}{3}{amod}
 \depedge{5}{4}{obj}
\wordgroup[style={blue}]{1}{1}{2}{none}
\wordgroup[style={red}]{1}{3}{4}{none}
\wordgroup[style={blue}]{1}{5}{5}{none}
\end{dependency}}

\centering
\resizebox{.9\linewidth}{!}{
\begin{dependency}[label theme = simple, edge theme = iron, label style={font=\small}]
 \begin{deptext}[column sep=1em]
  Ram \& Kapoor \& reminds \& me \& of \& boondi \& ke \& laddu \\
 \end{deptext}
 \deproot[edge unit distance=3.25ex]{3}{ROOT}
 \depedge{3}{1}{nsubj}
 \depedge{1}{2}{flat}
 \depedge{3}{4}{obj}
 \depedge{8}{5}{case}
 \depedge{8}{6}{nmod}
 \depedge{6}{7}{case}
 \depedge[edge unit distance=2.25ex]{3}{8}{obl}
\wordgroup[style={black}]{1}{1}{2}{none}
\wordgroup[style={red}]{1}{3}{5}{none}
\wordgroup[style={blue}]{1}{6}{8}{none}
\end{dependency}}

\centering
\resizebox{.85\linewidth}{!}{
\begin{dependency}[label theme = simple, edge theme = iron, label style={font=\small}]
 \begin{deptext}[column sep=1em]
  Has \& someone \& told \& Gabbar \& cal \& kya \& hai \& ? \\
 \end{deptext}
 \deproot{3}{ROOT}
 \depedge{3}{1}{aux}
 \depedge{3}{2}{nsubj}
 \depedge{3}{4}{iobj}
 \depedge{6}{5}{nmod}
 \depedge[edge unit distance=2ex]{3}{6}{ccomp}
 \depedge{6}{7}{cop}
 \depedge[edge unit distance=1.75ex]{3}{8}{punct}
\wordgroup[style={red}]{1}{1}{3}{none}
\wordgroup[style={black}]{1}{4}{4}{none}
\wordgroup[style={blue}]{1}{5}{7}{none}
\end{dependency}}

%\resizebox{.7\linewidth}{!}{
%\begin{dependency}[label theme = simple, edge theme = iron, label style={font=\small}]
% \begin{deptext}[column sep=1em]
%  Manhoos \& train \& as \& usual \& late \& hai \& :( \\
% \end{deptext}
% \deproot[edge unit distance=2.25ex]{5}{ROOT}
% \depedge{2}{1}{amod}
% \depedge[edge unit distance=2ex]{5}{2}{nsubj}
% \depedge{4}{3}{case}
% \depedge{5}{4}{amod}
% \depedge{5}{6}{cop}
% \depedge{5}{7}{discourse}
%\end{dependency}}

%\resizebox{.7\linewidth}{!}{
%\begin{dependency}[label theme = simple, edge theme = iron, label style={font=\small}]
% \begin{deptext}[column sep=1em]
%  Tumhara \& happy \& birthday \& to \& you \& kab \& aata \& hai \& ? \\
% \end{deptext}
% \deproot{7}{ROOT}
% \depedge{3}{1}{nmod}
% \depedge{3}{2}{amod}
% \depedge[edge unit distance=2.25ex]{7}{3}{nsubj}
% \depedge{5}{4}{case}
% \depedge{3}{5}{nmod}
% \depedge{7}{6}{nmod}
% \depedge{7}{8}{aux}
% \depedge{7}{9}{punct}
%\end{dependency}}

\centering
\resizebox{.9\linewidth}{!}{
\begin{dependency}[label theme = simple, edge theme = iron, label style={font=\small}]
 \begin{deptext}[column sep=1em]
  Enjoying \& Dilli \& ki \& sardi \& after \& a \& long \& time \& . \\
 \end{deptext}
 \deproot[edge unit distance=3.85ex]{1}{ROOT}
 \depedge{4}{2}{nmod}
 \depedge{2}{3}{case}
 \depedge{1}{4}{obj}
 \depedge{8}{5}{case}
 \depedge{8}{6}{det}
 \depedge{8}{7}{amod}
 \depedge[edge unit distance=1.65ex]{1}{8}{obl}
 \depedge[edge unit distance=1.75ex]{1}{9}{punct}
\wordgroup[style={red}]{1}{1}{1}{none}
\wordgroup[style={black}]{1}{2}{2}{none}
\wordgroup[style={blue}]{1}{3}{4}{none}
\wordgroup[style={red}]{1}{5}{8}{none}
\end{dependency}}

%\centering
%\resizebox{1\linewidth}{!}{
%\begin{dependency}[label theme = simple, edge theme = iron, label style={font=\small}]
% \begin{deptext}[column sep=1em]
%  Our \& life \& revolves \& around \& LOG \& KYA \& KAHENGAY \& . \\
% \end{deptext}
% \deproot[edge unit distance=4ex]{3}{ROOT}
% \depedge{2}{1}{nmod}
% \depedge{3}{2}{nsubj}
% \depedge{7}{4}{mark}
% \depedge{7}{5}{nsubj}
% \depedge{7}{6}{obj}
% \depedge{3}{7}{advcl}
% \depedge{3}{8}{punct}
%\wordgroup[style={red}]{1}{1}{4}{none}
%\wordgroup[style={blue}]{1}{5}{7}{none}
%\end{dependency}}

%\resizebox{.7\linewidth}{!}{
%\begin{dependency}[label theme = simple, edge theme = iron, label style={font=\small}]
% \begin{deptext}[column sep=1em]
%  Aa \& bel \& mujhe \& maar \& is \& the \& new \& black \& . \\
% \end{deptext}
% \deproot[edge unit distance=3.5ex]{8}{ROOT}
% \depedge[edge unit distance=1.75ex]{8}{1}{csubj}
% \depedge{1}{2}{nsubj}
% \depedge{4}{3}{iobj}
% \depedge{2}{4}{parataxis}
% \depedge{8}{5}{cop}
% \depedge{8}{6}{det}
% \depedge{8}{7}{amod}
% \depedge{8}{9}{punct}
%\end{dependency}}

\centering
\resizebox{.9\linewidth}{!}{
\begin{dependency}[label theme = simple, edge theme = iron, label style={font=\small}]
 \begin{deptext}[column sep=1em]
  Biggboss \& dekhne \& wali \& awaam \& can \& unfollow \& me \& . \\
 \end{deptext}
 \deproot[edge unit distance=2ex]{6}{ROOT}
 \depedge{2}{1}{obj}
 \depedge{4}{2}{amod}
 \depedge{2}{3}{mark}
 \depedge{6}{4}{nsubj}
 \depedge{6}{5}{aux}
 \depedge{6}{7}{iobj}
 \depedge{6}{8}{punct}
\wordgroup[style={black}]{1}{1}{1}{none}
\wordgroup[style={blue}]{1}{2}{4}{none}
\wordgroup[style={red}]{1}{5}{7}{none}
\end{dependency}}

\centering
\resizebox{.6\linewidth}{!}{
\begin{dependency}[label theme = simple, edge theme = iron, label style={font=\small}]
 \begin{deptext}[column sep=1em]
  Kaafi \& depressing \& situation \& hai \& yar \\
 \end{deptext}
 \deproot[edge unit distance=2ex]{3}{ROOT}
 \depedge{2}{1}{advmod}
 \depedge{3}{2}{amod}
 \depedge{3}{4}{cop}
 \depedge{3}{5}{vocative}
\wordgroup[style={blue}]{1}{1}{1}{none}
\wordgroup[style={red}]{1}{2}{3}{none}
\wordgroup[style={blue}]{1}{4}{5}{none}
\end{dependency}}

%\resizebox{.7\linewidth}{!}{
%\begin{dependency}[label theme = simple, edge theme = iron, label style={font=\small}]
% \begin{deptext}[column sep=1em]
%  Qul \& mila \& k \& I \& 've \& got \& an \& hour \& . \\
% \end{deptext}
% \deproot{6}{ROOT}
% \depedge{2}{1}{compound}
% \depedge[edge unit distance=2ex]{6}{2}{advcl}
% \depedge{2}{3}{aux}
% \depedge{6}{4}{nsubj}
% \depedge{6}{5}{aux}
% \depedge{8}{7}{det}
% \depedge{6}{8}{nmod}
% \depedge{6}{9}{punct}
%\end{dependency}}

\centering
\resizebox{.85\linewidth}{!}{
\begin{dependency}[label theme = simple, edge theme = iron, label style={font=\small}]
 \begin{deptext}[column sep=1em]
  Some \& people \& are \& double \& standards \& ki \& dukaan \\
 \end{deptext}
 \deproot{7}{ROOT}
 \depedge{2}{1}{det}
 \depedge[edge unit distance=2ex]{7}{2}{nsubj}
 \depedge[edge unit distance=2ex]{7}{3}{cop}
 \depedge{5}{4}{amod}
 \depedge{7}{5}{nmod}
 \depedge{5}{6}{case}
\wordgroup[style={red}]{1}{1}{5}{none}
\wordgroup[style={blue}]{1}{6}{7}{none}
\end{dependency}}

\centering
\resizebox{1\linewidth}{!}{
\begin{dependency}[label theme = simple, edge theme = iron, label style={font=\small}]
 \begin{deptext}[column sep=1em]
  There \& is \& no \& seperate \& emoji \& for \& khushi \& ke \& aansu \& . \\
 \end{deptext}
 \deproot[edge unit distance=4ex]{5}{ROOT}
 \depedge{5}{1}{expl}
 \depedge{5}{2}{cop}
 \depedge{5}{3}{advmod}
 \depedge{5}{4}{amod}
 \depedge{9}{6}{case}
 \depedge{9}{7}{nmod}
 \depedge{7}{8}{case}
 \depedge{5}{9}{obl}
 \depedge{5}{10}{punct}
\wordgroup[style={red}]{1}{1}{6}{none}
\wordgroup[style={blue}]{1}{7}{9}{none}
\end{dependency}}

%\end{document}